# QUATERNION ORTHOGONAL TRANSFORMER FOR FACIAL EXPRESSION RECOGNITION IN THE WILD

*Yu Zhou[1†], Liyuan Guo[2†], Lianghai Jin[1]\**

School of Computer Science and Technology[1], Institute of Artificial Intelligence[2],
Huazhong University of Science and Technology, China

## ABSTRACT

Facial expression recognition (FER) is a challenging topic in artificial intelligence. Recently, many researchers have attempted to introduce Vision Transformer (ViT) to the FER task. However, ViT cannot fully utilize emotional features extracted from raw images and requires a lot of computing resources. To overcome these problems, we propose a quaternion orthogonal transformer (QOT) for FER. Firstly, to reduce redundancy among features extracted from pre-trained ResNet-50, we use the orthogonal loss to decompose and compact these features into three sets of orthogonal sub-features. Secondly, three orthogonal sub-features are integrated into a quaternion matrix, which maintains the correlations between different orthogonal components. Finally, we develop a quaternion vision transformer (Q-ViT) for feature classification. The Q-ViT adopts quaternion operations instead of the original operations in ViT, which improves the final accuracies with fewer parameters. Experimental results on three in-the-wild FER datasets show that the proposed QOT outperforms several state-of-the-art models and reduces the computations.
Codes are available at https://github.com/Gabrella/QOT.

## 1. INTRODUCTION

In recent years, facial expression recognition (FER) has drawn much attention in artificial intelligence and computer vision. Improving the FER performance contributes to various applications such as psychological treatment and human-machine interaction. However, the current facial expression images are mainly collected from in-the-wild scenarios, which brings more challenges to FER than in-the-lab. Besides, many related works [1-2] have extracted features by pre-trained backbones. However, these features usually contain much redundant information, which may have a negative effect on the final performance.

To reduce the redundancy among features, Lin *et al*. [3] and Chen *et al*. [4] employed orthogonal loss to decompose the features into two orthogonal sub-features for multi-task. Inspired by their works, we develop the orthogonal feature decomposition to map the extracted features into three sets of orthogonal sub-features.

Moreover, some works [5-6] have shown that quaternion theory can explore internal correlations between different components and reduce the parameters. Gaudet *et al*. [7] and Zhou *et al*. [5] applied quaternion operations to deal with the multichannel images. However, these works [5-7] all focused on the correlations between RGB channels. To the best of our knowledge, we are the first work integrating orthogonal features into quaternion representation. This strategy can establish the transformation relationship among three orthogonal features as the related works [5-7]. Further, quaternion operations such as quaternion fully-connected layer and quaternion convolutional layer are able to process the orthogonal features holistically rather than separate them as independent components. The proposed quaternion orthogonal representation encodes the sub-features into a more efficient form for further processing.

In 2021, Dosovitskiy *et al*. [8] introduced a vision transformer (ViT) for image recognition and achieved better performance than CNN. Since then, the ViT has been widely used to handle computer vision tasks. Many works [1], [9] also made efforts to apply ViT to perform the FER. However, these methods used the original ViT structure directly, which takes up lots of computing resources and needs to be further improved. To this end, we propose a quaternion vision transformer (Q-ViT) to categorize the quaternion orthogonal features into different emotions, which effectively improves the final accuracy and reduces the parameters and floating-point operations per second (FLOPs) compared to ViT.

In this work, we propose a quaternion orthogonal transformer (QOT) for FER. It is evaluated on three in-the-wild FER datasets and obtains competitive performance. Our contributions are summarized as follows: *(i)* We propose an orthogonal feature decomposition to map the extracted features to three orthogonal sub-features. *(ii)* We propose the quaternion orthogonal representation, which correlates the orthogonal sub-features by quaternion theory. *(iii)* We develop the Q-ViT, which applies quaternion operations to deal with quaternion features and reduces the computations.

† Equal contributions
\* Corresponding author

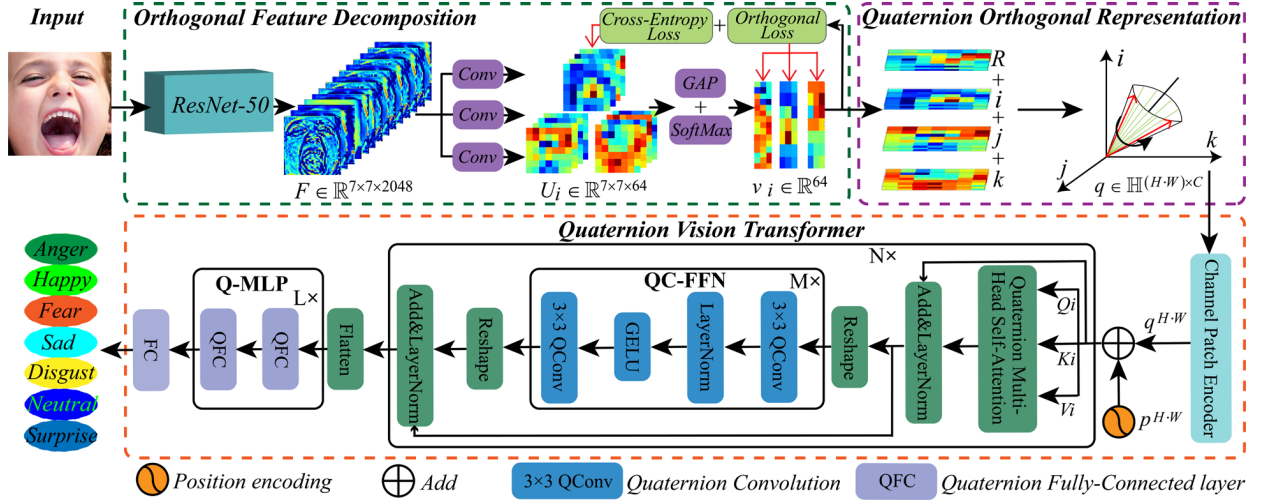

Fig. 1. The overview of the proposed QOT for FER.

## 2. QUATERNION ORTHOGONAL TRANSFORMER

### 2.1. Framework overview

**Fig. 1** shows the overview of the proposed QOT. The original images are fed into the Orthogonal Feature Decomposition module to obtain three sets of orthogonal sub-features. Then, three sets of orthogonal sub-features are integrated to quaternion matrix by Quaternion Orthogonal Representation. Finally, the Quaternion Vision Transformer processes these quaternion features and outputs the final emotional result.

### 2.2. Orthogonal Feature Decomposition

To reduce the redundancy among features, we develop the orthogonal feature decomposition to map the extracted features into three orthogonal sub-features. As shown in **Fig. 1**, we first choose the pre-trained ResNet-50 as the backbone to extract emotional feature $F \in \mathbb{R}^{7 \times 7 \times 2048}$ from a raw image. Then, we add three individual 1×1 convolutional layers at the end of the backbone to generate three compact features $U_i \in \mathbb{R}^{7 \times 7 \times 64}$, $i = 1, 2, 3$. Next, we apply the Global Average Pooling (GAP) and SoftMax operations to transfer $U_i$ to vectors $v_i \in \mathbb{R}^{64}$. Finally, we introduce an orthogonal loss $L_{ortho}$ to force the intermediate features $U_i$ and $v_i$ to be orthogonal. Moreover, the Cross-Entropy loss is combined with the orthogonal loss to fine-tune the modified backbone. The orthogonal loss and the combined loss are expressed as follows:

$$L_{ortho} = \sum \left| \frac{v_i \cdot v_j}{\|v_i\|_2 \cdot \|v_j\|_2} \right| / n, \quad i, j \in \{1, 2, 3\} \quad (1)$$

$$L = L_{CrossEntropy} + \lambda L_{ortho} \quad (2)$$

where $n$ is the number of vectors. $|\cdot|$ denotes the absolute value operator and $\|\cdot\|_2$ represents the L2 norm. $\lambda$ is the hyper-parameter that balances the two loss functions.

### 2.3. Quaternion Orthogonal Representation

Quaternion representation has been wildly used in color image processing. The existing works usually put RGB channels of an image into the imaginary parts of the quaternion matrix, which is expressed as follows:

$$Q(x, y) = R(x, y)i + G(x, y)j + B(x, y)k \quad (3)$$

where $R(x, y)$, $G(x, y)$ and $B(x, y)$ are the red, green, and blue channel values, respectively.

However, few works try to integrate orthogonal features into quaternion representation. In our work, we develop the quaternion orthogonal representation to maintain the internal dependencies between different orthogonal features. To be specific, three orthogonal features are assigned to three imaginary parts of the quaternion matrix, respectively. Moreover, to balance three decomposed orthogonal features, the average features of orthogonal features are set as the real part of the quaternion matrix. Therefore, the quaternion orthogonal representation can be expressed as

$$Q(x, y) = Ave(x, y) + f_1(x, y)i + f_2(x, y)j + f_3(x, y)k \quad (4)$$

where $f_1(x, y)$, $f_2(x, y)$ and $f_3(x, y)$ are three orthogonal features. $Ave(x, y)$ is the average feature of them. The proposed quaternion orthogonal representation aims to correlate the orthogonal features and make the extracted information more efficient.

### 2.4. Quaternion Vision Transformer

In this work, we extend the ViT to the quaternion domain and develop a quaternion vision transformer (Q-ViT) for classification. As shown in **Fig. 1**, the overall framework of Q-ViT follows that of the ViT and it makes several crucial improvements on the key components. These improvements include channel patch encoder, quaternion multi-head self-attention (Q-MHSA) and quaternion convolution feed-forward network (QC-FFN).

Different from the patch encoder in ViT, channel patch encoder splits input features into patches along the channel axis. For instance, the quaternion feature $q \in \mathbb{H}^{H \times W \times C}$ is first reshaped to 2D patch $q \in \mathbb{H}^{(H \cdot W) \times C}$ and then encoded to $C$ number of flattening sequences $q^{H \cdot W}$, where $H$ and $W$ are the height and width of the feature map, $H \cdot W$ is the sequence length and $C$ is the channel number. In this work, $H$ and $W$ are set to 7 and $C$ is set to 64. Moreover, to retain the channel positional information, 1D position embedding $p^{H \cdot W}$ is added to $q^{H \cdot W}$. It is worth noting that $q^{H \cdot W}$ is a quaternion sequence and $p^{H \cdot W}$ is also a quaternion number. Their output sequences are finally fed into the Q-MHSA.

Q-MHSA is produced with quaternion operations, which handle quaternion features with a quarter of parameters compared to ordinary operations of ViT [8]. The quaternion operations mainly contain quaternion fully-connected layer $QFC(\cdot)$ and quaternion convolution layer $QConv(\cdot)$. The fully-connected layer is seen as a special one-dimensional convolutional layer. Both of them follow the rule of quaternion multiplication. For instance, the quaternion matrixes $Q$ is convolved with the quaternion kernel $g$ can be formulated as:

$$Q \otimes g = \begin{bmatrix} 1 & i & j & k \end{bmatrix} \cdot \left( \begin{bmatrix} Q_r & -Q_i & -Q_j & -Q_k \\ Q_i & Q_r & -Q_k & Q_j \\ Q_j & Q_k & Q_r & -Q_i \\ Q_k & -Q_j & Q_i & Q_r \end{bmatrix} \otimes \begin{bmatrix} g_r \\ g_i \\ g_j \\ g_k \end{bmatrix} \right) \quad (5)$$

where $\otimes$ is the Hamilton product.

In Q-MHSA, we first utilize three individual $QFC(\cdot)$ to transform quaternion sequence $q$ to quaternion query $Q$, quaternion key $K$ and quaternion value $V$ as **Eq. (6)**. Then, the self-attention is computed by Hamilton product instead of scaled dot product in ViT, which is shown as **Eq. (7)**. The *ComponentSoftmax* operation means that Softmax operation on each component of the quaternion. Next, the self-attentions are concatenated and mapped to a multi-head quaternion matrix by $QFC(\cdot)$ as **Eq. (8)**. Finally, the multi-head quaternion features from the Q-MHSA are sent to QC-FFN for further processing. The whole computing process of Q-MHSA can be formulated as:

$$Q = QFC_q(q); K = QFC_k(q); V = QFC_v(q) \quad (6)$$

$$\begin{aligned} head_j &= Attention(Q,K,V) \\ &= ComponentSoftmax\left(\left(Q \otimes K^T\right)/\sqrt{d}\right) \otimes V \end{aligned} \quad (7)$$

$$MultiHead = QFC\left(Concat(head_1, \ldots, head_j)\right), j = 8 \quad (8)$$

where $Concat(\cdot)$ represents the concatenation operation.

To better capture local features, we also introduce the $QConv(\cdot)$ to QC-FFN. To be specific, the quaternion features are normalized by layer normalization and then fed into the QC-FFN. The QC-FFN consists of quaternion convolution, GELU activation and layer normalization. The computation of QC-FFN is as follows:

$$X = LayerNorm(MultiHead) \quad (9)$$

$$\begin{aligned} X_{out} &= QC\text{-}FFN(X) \\ &= QConv(GELU(LayerNorm(QConv(X)))) \end{aligned} \quad (10)$$

where *LayerNorm* represents layer normalization, *QConv* is quaternion convolution and *GELU* [10] is activation operation. $X$ and $X_{out}$ are input and output of QC-FFN.

In the classification part, the quaternion matrix $X_{out}$ is reshaped and passed through the *Add & LayerNorm*. Next, the quaternion features are flattened into a 1D vector. It is connected to quaternion multi-layer perception (Q-MLP) and a fully-connected layer $FC(\cdot)$ for final classification.

## 3. EXPERIMENTS

### 3.1. Dataset

**SFEW** dataset [11] is comprised of static frames from different movies, which has been divided into training set (958 samples), validation set (436 samples) and test set (372 samples). Because the labels of test set are not released, we evaluate the models on the validation set as other works. **RAF-DB** dataset [12] is comprised of facial expression images from the Internet. In our work, we choose the seven basic emotions including 12271 training samples and 3068 test samples to evaluate the model. **AffectNet** dataset [13] contains 450000 annotated facial expression images. In this work, we only use the seven basic emotions including 28000 training samples and 3500 test samples. Each emotion has 4000 training samples and 500 testing samples.

### 3.2. Implementation Details

The QOT is implemented by Tensorflow on Python 3.8 platform, using a computer server running Linux with Intel Core i7 3.4GHz CPU with an NVIDIA GeForce GTX 1080Ti GPU. First, the ResNet-50 pre-trained on the MS-Celeb-1M dataset is used to be the backbone and three individual convolutional layers are added at the end of this backbone. Next, the orthogonal loss and Cross-Entropy loss are combined to fine-tune the modified backbone and obtain three sets of orthogonal sub-features. These orthogonal features and their average feature are then integrated into a quaternion matrix. Finally, the quaternion matrixes are fed into Q-ViT. There are many hyperparameters in Q-ViT that need to be set. In this work, N in Q-ViT is set to 4, M in QC-FFN is set to 2 and L in Q-MLP is set to 2.

## 3.3. Visualization Results

To intuitively show the performance of QOT, we utilize the t-SNE method to visualize the experimental results. **Fig. 2** visualizes the latent features produced by the baseline ResNet-50 (row I) and QOT (row II). The different color dots represent different emotions. Obviously, QOT is more capable of separating the data into seven compact clusters than the baseline. Moreover, we also apply the confusion matrix to show the recognition results of each expression. **Fig. 3** shows that QOT can predict most expressions except a few disgust and fear expressions.

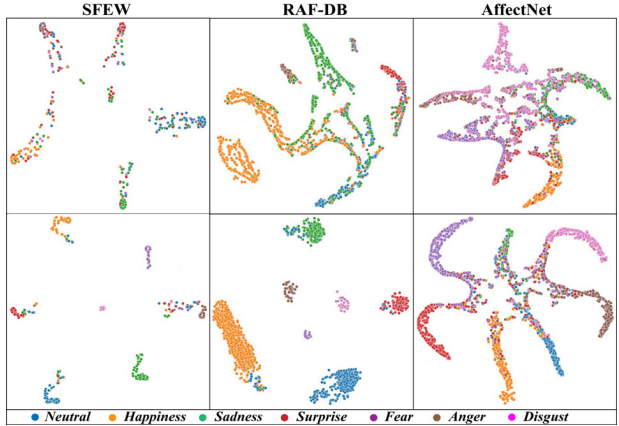

Fig. 2. T-SNE visualization results of baseline (row I) and QOT (row II).

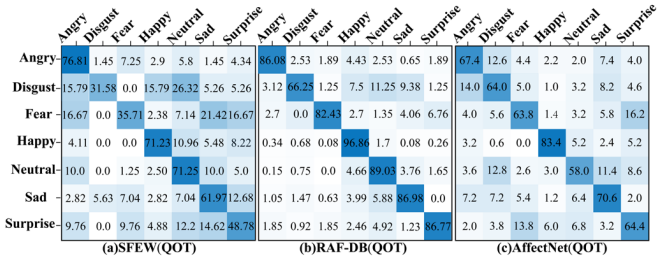

Fig. 3. Confusion matrixes of QOT on three datasets.

## 3.4. Experimental Results

To verify the effectiveness of QOT, we conduct ablation studies of the quaternion module, orthogonal features and ViT module, respectively. The accuracies, parameters and FLOPs of comparative models are shown in **Table I**. First, we remove the quaternion module from QOT and construct an Ortho-ViT for ablation experiments. The comparative results show that quaternion module in QOT contributes to improving the accuracies by 1.83%, 0.89% and 0.11% on SFEW, RAF-DB and AffectNet, respectively. Moreover, quaternion modules in QOT also reduce the Params and FLOPs compared to Ortho-ViT. Then, we remove orthogonal features from QOT and construct Q-ViT for comparison. From the results, orthogonal features contribute to obtaining 27.39%, 14.12% and 14.83% higher accuracies on three datasets. Finally, we remove the ViT module from QOT and construct Ortho-CNN for comparison. Without the ViT module, the accuracies drop by 4.81%, 1.03% and 1.89% on three datasets. Generally, the three modules all contribute greatly to the QOT.

TABLE I. THE ABLATION STUDIES OF QOT.

| Method | SFEW | RAF-DB | AffectNet | Params | FLOPs |
|---|---|---|---|---|---|
| Q-ViT | 35.14% | 75.85% | 52.54% | 7.04M | 10.53M |
| Ortho-CNN | 57.72% | 88.94% | 65.48% | 23.90M | 4.15G |
| Ortho-ViT | 60.70% | 89.08% | 67.26% | 21.59M | 21.58M |
| **QOT** | **62.53%** | **89.97%** | **67.37%** | **7.04M** | **10.53M** |

We also compare QOT with several state-of-the-art methods on three datasets. The results are shown in **Table II**, which demonstrates that QOT generally achieves higher accuracies with fewer computation resources. Specifically, compared with FDRL [16] on SFEW, QOT achieves 0.37% higher accuracy. Although QOT doesn't reach the minimum number of parameters, it still reduces the number of FLOPs and obtains the highest accuracy. Compared with VTFF [1], which also applies a ViT with CNN features, QOT obtains 1.83% and 5.52% higher accuracies on RAF-DB and AffectNet. Compared with MA-Net [14], QOT gets 1.57% and 2.84% higher accuracy on RAF-DB and AffectNet. The comparison results demonstrate that QOT is cost-efficient both in performance and computation complexity.

TABLE II. ACCURACIES OF QOT AND OTHER STATE-OF-THE-ART METHODS

| Datasets | Methods | Year | Accuracy | Params | FLOPs |
|---|---|---|---|---|---|
| SFEW | ViT [8] | 2021 | 32.91% | 21.59M | 21.58M |
|  | Baseline | 2016 | 52.05% | 23.52M | 4.13G |
|  | MA-Net [14] | 2021 | 59.40% | 50.54M | 3.65G |
|  | CS-GResNet [15] | 2022 | 60.55% | **2.80M** | 1.72G |
|  | FDRL [16] | 2021 | 62.16% | - | - |
|  | **QOT** | 2022 | **62.53%** | 7.04M | 10.53M |
| RAF-DB | ViT [8] | 2021 | 74.76% | 21.59M | 21.58M |
|  | Baseline | 2016 | 87.39% | 23.52M | 4.13G |
|  | MAPNet [17] | 2022 | 87.26% | - | - |
|  | VTFF [1] | 2022 | 88.14% | 51.8M | - |
|  | MA-Net [14] | 2021 | 88.40% | 50.54M | 3.65G |
|  | **QOT** | 2022 | **89.97%** | 7.04M | 10.53M |
| AffectNet | ViT [8] | 2021 | 51.14% | 21.59M | 21.58M |
|  | Baseline | 2016 | 64.37% | 23.52M | 4.13G |
|  | VTFF [1] | 2022 | 61.85% | 51.8M | - |
|  | MAPNet [17] | 2022 | 64.09% | - | - |
|  | MA-Net [14] | 2021 | 64.53% | 50.54M | 3.65G |
|  | **QOT** | 2022 | **67.37%** | 7.04M | 10.53M |

## 4. CONCLUSION

This paper proposes a quaternion orthogonal transformer (QOT) for FER in-the-wild. It proposes the orthogonal feature decomposition to extract features and decompose them into three orthogonal sub-features. Then, the quaternion orthogonal representation correlates the orthogonal features by quaternion theory. Finally, Q-ViT classifies quaternion features into emotions and reduces the computations. The QOT has achieved the SOTA accuracy with 62.53%, 89.97% and 67.37% on SFEW, RAF-DB and AffectNet, respectively.